\documentclass[12pt]{article}

\usepackage[margin=1in]{geometry}
\usepackage[T1]{fontenc}
\usepackage[utf8]{inputenc}
\usepackage{textcomp}
\usepackage{lmodern}
\usepackage{microtype}

\usepackage{amsmath,amssymb,amsthm}
\usepackage{booktabs}
\usepackage{array}
\usepackage{longtable}

\usepackage{graphicx}
\usepackage{caption}
\usepackage{float}
\usepackage{xcolor}
\usepackage{xurl}
\usepackage{natbib}
\usepackage{hyperref}

\hypersetup{
  pdftitle={Sequential Statistical Inference for Large Language Models: Representation, Validity, and Monitoring},
  pdfauthor={Yao Xie},
  pdfkeywords={large language models, sequential inference, conformal prediction, change-point detection, online monitoring, dependent data},
  colorlinks=true,
  linkcolor=blue,
  citecolor=blue,
  urlcolor=blue
}

\title{Sequential Statistical Inference for Large Language Models:
Representation, Validity, and Monitoring}
\author{
Yao Xie\thanks{This article was prepared for a invited discussion in \textit{The American Statistician}.}\\
H. Milton Stewart School of Industrial and Systems Engineering\\
Georgia Institute of Technology
}
\date{\today}

\begin{document}

\maketitle

\begin{abstract}
This discussion argues that sequential statistical inference can naturally contribute to LLM trustworthiness. In deployment, LLM systems are queried repeatedly, conditioned on evolving contexts, and incorporate user or tool feedback, and may exhibit behavioral shifts after model updates or distribution changes. The discussion is organized around three tasks: \emph{representation}, modeling LLM interactions as dependent stochastic processes rather than isolated prompt--response pairs; \emph{validity}, developing uncertainty guarantees that remain meaningful under dependence, repeated use, and adaptation; and \emph{monitoring}, using sequential alarms and change-point detection to identify shifts in calibration, hallucination rates, refusal behavior, fairness, or other task-relevant properties. This perspective complements recent surveys by viewing trustworthy LLM deployment as a problem of statistical process control.

\end{abstract}


\section{Introduction}\label{sec:intro}

The overview by \citet{ji2026overview} makes a strong case that statisticians have an important role to play in LLM research, especially through uncertainty quantification, conformal prediction, interpretability, fairness, privacy, and human--AI collaborative data science. This discussion builds on that agenda by focusing on one statistical structure that appears across many of these themes: sequential inference.

Large language models (LLMs) are often introduced through the lens of next-token prediction. This is natural and mathematically convenient, but it gives only a partial statistical description of how these systems behave in practice. Once deployed, LLMs are queried repeatedly, conditioned on evolving contexts, interact with users and tools, and influence future prompts and decisions. Their outputs are therefore better viewed not as isolated predictions, but as parts of a dependent and adaptive stochastic process.

The perspective developed here is shaped by sequential inference, change-point detection, and conformal inference for dependent data. From this viewpoint, trustworthy LLM systems should be treated not only as a problem of static uncertainty quantification but also as one of online validity, adaptive feedback, and post-deployment monitoring.

This discussion advances a simple position: an important statistical layer for LLMs is a \emph{sequential inference} layer. This layer has three connected components. First, LLM behavior should be represented at the level of dependent interaction trajectories, rather than isolated prompt--response pairs. Second, reliability should be framed in terms of validity guarantees that remain meaningful under dependence, repeated use, and adaptive interaction. Third, deployed systems should be monitored over time, with sequential alarms and change-point detection used to identify abrupt shifts in calibration, safety, refusal behavior, or other task-relevant properties.

Put differently, the sequential view changes the unit of analysis from a single prompt--response pair to an interaction trajectory, the reliability target from confidence in one answer to online validity under dependence, and the deployment question from ``Is this answer good?'' to ``Has the system's behavior changed?'' This framing is intended to complement the broad overview of \citet{ji2026overview}, not to replace it. The sequential perspective is not separate from their agenda; rather, it makes explicit a common structure that runs through several topics in their overview, including uncertainty quantification, conformal prediction, model adaptation, fairness, and human--AI collaboration.

The remainder of this discussion develops the three components of this position. Section~\ref{sec:representation} argues for a sequential statistical representation of LLM behavior. Section~\ref{sec:validity} reframes uncertainty as an online validity problem under dependence. Section~\ref{sec:monitoring} argues that monitoring and change-point detection should become central tools for deployed LLM systems.

\section{Representation: A Sequential View}\label{sec:representation}

The usual account of an LLM begins with a sequence of tokens $x_1,x_2,\dots,x_T$ and the conditional distribution of the next token given the past. From the perspective of language modeling, this is the correct primitive. From the perspective of statistical deployment, however, it is often not sufficient. In practical use, the objects of interest are not merely token sequences, but full interaction histories involving prompts, retrieved context, observable intermediate outputs, tool outputs, user feedback, and future actions.

A useful statistical abstraction is to view an LLM-driven workflow as a dependent process. Suppose that each completed interaction is recorded as \(X_t=(P_t,R_t,U_t)\), where \(P_t\) is the prompt or context at stage \(t\), \(R_t\) is the model response, and \(U_t\) is any user, tool, or environmental feedback observed after the response and potentially incorporated into future stages. The interaction history and associated filtration are then \(\mathcal{H}_t=(X_1,\ldots,X_t)\) and \(\mathcal{F}_t=\sigma(X_1,\ldots,X_t)\), with \(\mathcal{F}_0\) denoting the initial information available before deployment or before the start of the interaction stream. At stage \(t\), the prompt or context \(P_t\) may itself be chosen adaptively based on \(\mathcal{F}_{t-1}\). The response is therefore better regarded through its conditional law \(R_t \mid (P_t,\mathcal{F}_{t-1}) \sim Q_t(\cdot\mid P_t,\mathcal{F}_{t-1})\), rather than as an isolated draw. The kernel \(Q_t\) may reflect the model, decoding rule, retrieval system, deployment regime, and so on.

This reframing matters for several reasons. It clarifies the dependence structure: multi-turn prompting, retrieval augmentation, tool use, and interactive editing all create histories for which exchangeability is difficult to justify. It also changes what should be predicted or validated. In many settings, the object of interest is not the probability of a token, but the reliability of a trajectory-level output, such as a recommendation, a diagnosis, a summary, or a sequence of actions. Finally, it creates a natural interface with the language of stochastic processes, filtrations, online inference, and state evolution.

This viewpoint also places LLMs in continuity with a broader class of dependent-sequence models familiar to statisticians. For instance, time series, spatio-temporal processes, point processes, state space models, and sequential generalized linear models are all examples of systems in which dependence, partial observation, and evolving context matter. Recent work on sequential generalized linear and nonlinear time-series models provides one related statistical perspective \citep{JuditskyEtAl2023,ZhouXie2025}.  One can regard next-token prediction as a particular instance of sequence modeling, although one with an unusually rich and high-dimensional state. 

A related issue is the choice of representation. One can distinguish between the observable interaction history available to the analyst and the latent computational representation used internally by the model. For statistical inference, the observable history is the more natural object: it is the record on which validity, monitoring, and auditing can be based. Internal representations may provide useful additional information when available, for example, for diagnostics or uncertainty scores, but they are typically model-specific and may not be accessible in deployed systems. This suggests that statistical procedures for LLM deployment should be formulated first at the level of observable trajectories, with internal representations used only as optional refinements.

This representation also clarifies what can be monitored. From each completed interaction, one may define a task-specific score \(S_t=\phi_t(X_t,\mathcal{F}_{t-1})\), with \(S_t\in\mathcal{F}_t\), where \(S_t\) summarizes a statistical property of interest for the deployed system, such as reliability, validity, safety, or alignment with external feedback. The sequence \(\{S_t\}_{t\ge 1}\) is then a lower-dimensional statistical summary of the interaction process, providing a bridge between rich LLM trajectories and classical tools for online inference and change-point detection.

In a nutshell, for statisticians, an LLM should not be viewed solely as a one-step probabilistic predictor. It is more naturally represented as a dependent stochastic process evolving over context, interaction, and feedback. Without this representational move, the later questions of validity and monitoring are harder to pose cleanly.

\section{Validity: Sequential Uncertainty Quantification}
\label{sec:validity}

A central theme in \citet{ji2026overview} is that statisticians can contribute to trustworthy LLMs through uncertainty quantification, conformal prediction, and related inferential tools. The sequential perspective sharpens this point: in deployment, uncertainty is not only about a single response, but about whether a statistical guarantee remains meaningful over repeated, dependent, and adaptive interactions.

It is useful to distinguish \emph{confidence} from \emph{validity}. Confidence is often a model-facing quantity, derived from token probabilities, entropy, self-consistency, or auxiliary scoring systems. Validity is a procedure-facing quantity: it asks whether an uncertainty statement has the advertised statistical interpretation under the deployment conditions of interest. For LLMs, this distinction matters because a response may appear confident while being poorly calibrated for factual correctness, task success, or downstream loss.

Conformal prediction is attractive because it gives externally checkable uncertainty guarantees. Its classical form, however, relies on exchangeability. This assumption is difficult to justify for deployed LLM systems: prompts arrive over time, users adapt to previous responses, retrieval sources change, tools create feedback loops, and model updates may shift the response distribution. This lack of exchangeability is not a reason to abandon conformal ideas. It is precisely the setting that motivates online conformal inference, conformal inference under distribution shift, and conformal prediction for dependent time-series data \citep{xu2021conformal,GibbsCandes2024,jiang2026leave}.

Using the notation of Section~\ref{sec:representation}, each interaction can be represented as $X_t=(P_t,R_t,U_t)$, adapted to the filtration $\mathcal{F}_t$. A task-specific score $S_t=\phi_t(X_t,\mathcal{F}_{t-1})$, with $S_t\in\mathcal{F}_t$, may summarize a property of interest, such as reliability, validity, safety, or agreement with external feedback. Depending on the application, $S_t$ may serve as a conformity score, calibration error, answer-quality score, or downstream loss.

The key point is that validity should be assessed over the stream $\{S_t\}_{t\ge 1}$, not only over isolated prompts. Calibration may need to account for local history, context, user population, retrieval source, model version, and deployment regime. Repeated prompting or selective use of model outputs also raises optional-stopping issues, making anytime-valid reasoning a natural fit \citep{HowardEtAl2021}.

Thus, the validity problem for LLMs has several levels: uncertainty for a single response, calibrated guarantees for a task-specific score, and validity over an adaptive interaction stream. The last level is often the most relevant in deployment. Once such a sequential score is constructed, it also becomes the basic signal for monitoring. In this sense, validity supplies the statistical quantity on which online assessment and change detection can operate.

\section{Monitoring via Online Change-Point Detection}
\label{sec:monitoring}

The discussion of trustworthy LLMs in \citet{ji2026overview} raises a post-deployment question. Even if an LLM system begins with acceptable calibration, uncertainty quantification, fairness behavior, and task performance, these properties may not remain stable after deployment. Prompt distributions shift, user populations change, retrieval corpora evolve, model updates alter behavior, and adversarial usage patterns may appear. A statistical theory of LLM trustworthiness, therefore, needs not only pre-release evaluation, but also online monitoring.

The operational question is whether the system has entered a new regime. Suppose that each completed interaction produces a monitored score \(Z_t\), summarizing a property such as reliability, calibration, safety, agreement with external feedback, or task performance. The important point is not that there is one universal score, but that a rich LLM workflow can often be summarized by a time-indexed sequence \(\{Z_t\}_{t\ge 1}\) on which sequential testing can be performed.

An online monitoring procedure can be represented by an alarm time, mathematically a stopping time,
\(\tau=\inf\{t\geq 1:W_t(Z_1,\ldots,Z_t)>b\}\),
where \(W_t\) is a detection statistic and \(b\) is chosen to control false alarms under a reference regime. This formulation connects LLM monitoring to classical online change-point detection, where one seeks to detect a shift from a baseline distribution to a post-change distribution while controlling false alarms and minimizing detection delay \citep{Page1954,BassevilleNikiforov1993}. Modern procedures also account for memory use and computational cost in this tradeoff \citep{WangXie2024}. In the simplest model, \(Z_t\sim F_0\) before an unknown change time \(\kappa\) and \(Z_t\sim F_1\) after \(\kappa\). For LLM interaction streams, the more relevant formulation may be a change in the conditional distribution of \(Z_t\) given the past, with possible non-stationarity post-change, as in dependent-data formulations of sequential change-point detection \citep{TartakovskyEtAl2014}.

This perspective is related to the concept-drift literature in machine learning, but the LLM setting is broader. Traditional concept drift usually concerns changes in the data-generating distribution or in the relation between inputs and labels in an online prediction problem \citep{GamaEtAl2014}. In LLM deployment, the drift may occur not only in the prompt distribution but also in the model, retrieval system, tool environment, user feedback loop, or policy governing refusals and safety behavior. Such changes can alter calibration, retrieval reliability, refusal behavior, or unsafe-response patterns, making them better viewed as regime changes than as isolated errors. The monitored quantity may therefore be a calibration score, conformity score, human correction signal, retrieval-consistency score, or subgroup-specific error rate, rather than a standard supervised prediction loss.

A related line of work studies control charts for multi-agent systems, where fixed baselines may be inadequate when deployed agents learn, drift, or face adaptive inputs \citep{HelmPriebeDuderstadt2026}.  The broader statistical workflow is to define a monitored score linked to validity, safety, or task performance; estimate its reference behavior; and apply an online rule to detect persistent departures. When an alarm occurs, the system may trigger recalibration, fallback, auditing, human review, or suspension of a specific feature.

Monitoring also links back to validity. If online conformal or calibration procedures generate sequential scores, then change detection can be applied to those same quantities. Conversely, when a change-point detector signals a regime break, old calibration guarantees may need to be suspended and rebuilt under the new regime. Thus, validity and monitoring should be designed jointly: validity defines the statistical quantity of interest, and monitoring asks whether that quantity remains stable after deployment.

\section{Conclusion}\label{sec:conclusion}

This discussion has argued that one distinctive contribution of statistics to large language models is a sequential inference layer built around \emph{representation, validity, and monitoring}. Representation defines the object of analysis as an evolving interaction trajectory; validity asks for guarantees that remain meaningful under dependence, repeated use, and adaptation; and monitoring asks whether those guarantees remain credible after deployment. This view complements the agenda of \citet{ji2026overview} by suggesting that uncertainty quantification, conformal prediction, fairness, privacy, interpretability, and human--AI collaboration share a common operational structure: deployed LLMs are adaptive systems observed over time. The practical implication is that trustworthy LLM deployment requires trajectory-level modeling, uncertainty procedures valid under dependence, monitored reliability scores, and sequential alarms for regime changes, linked to human--AI workflows for recalibration, auditing, escalation, or human review. In this sense, the LLM era makes classical statistical questions about dependence, validity, and change newly urgent, and sequential inference provides a natural language for addressing them.



\end{document}